\documentclass[conference]{IEEEtran}
\IEEEoverridecommandlockouts
\usepackage{cite}
\usepackage{amsmath,amssymb,amsfonts}
\usepackage{algorithm}
\usepackage{algorithmic}
\usepackage{bm}
\usepackage{graphicx}
\usepackage{textcomp}
\usepackage{xcolor}
\usepackage{subcaption}
\usepackage{balance}

\newcommand{\new}[1]{#1}

\DeclareMathOperator*{\argmin}{arg\,min}

\DeclareMathOperator{\OCE}{OCE}
\DeclareMathOperator{\CVaR}{CVaR}
\DeclareMathOperator{\sCVaR}{sCVaR}
\DeclareMathOperator{\VaR}{VaR}

\DeclareMathOperator{\UCB}{UCB}
\DeclareMathOperator{\Var}{Var}

\title{Conformal Risk-Averse Decision Making\\ with Optimized Certainty Equivalent Risk Control}

\author{\IEEEauthorblockN{Amirmohammad Farzaneh and Osvaldo Simeone}
\IEEEauthorblockA{{Institute for Intelligent Networked Systems (INSI), Northeastern University London, London, UK}\\
\{a.farzaneh, o.simeone\}@nulondon.ac.uk}}

\begin{document}

\maketitle

\begin{abstract}
We study risk-averse decision making, in which an agent selects actions while being uncertain  about the true system state. The risk is measured via optimized certainty equivalent (OCE) metrics, which generalize  popular criteria such as mean-variance risk and conditional value-at-risk (CVaR). We characterize the optimal policy under known distributions,  and show that it reduces to a prediction set-based solution for the CVaR. This provides an operational interpretation of conformal prediction-type prediction sets. For unknown distributions, we develop a data-driven calibration strategy, based on a synthetic model for the likelihood and held-out calibration data, yielding high-probability control of the OCE risk. The approach is evaluated on two wireless beamforming settings.
\end{abstract}


\section{Introduction}
\label{sec:intro}

Decisions under uncertainty require not only accurate predictions, but also principled ways  to account for predictive uncertainty~\cite{Simeone2026}. In many systems, from wireless communications to autonomous driving, an agent observes features $X$, which generally leave the true state $Y$ of the system uncertain. Based on input $X$, the agent selects an action $a(X)$, and incurs a loss $\ell(a(X),Y)$. A risk-neutral agent minimizes the expected loss $\mathbb{E}[\ell(a(X), Y)]$, but this criterion does not account for the variability of outcomes,  downplaying the impact of tail events.

Risk measures provide a systematic way to incorporate risk aversion into decision making. Coherent and convex risk measures~\cite{Artzner1999,Follmer2016} formalize desirable properties such as subadditivity, i.e., diversification cannot increase risk. Optimized certainty equivalent (OCE) risks~\cite{BenTal1986,BenTal2007} form a broad class of convex risk measures that includes conditional value-at-risk (CVaR), entropic risk, and mean-variance risk as special cases. OCE risk measures admit efficient sample-based estimation~\cite{Gupte2026}, and can be formulated in terms of distributionally robust optimization~\cite{Rahimian2022}.

An active line of work has developed conformal prediction~\cite{Angelopoulos2022} as a framework for distribution-free uncertainty quantification. Reference~\cite{Kiyani2025} recently established decision-theoretic foundations for conformal prediction, showing that a risk-averse agent using a relaxation of the value-at-risk (VaR) risk, i.e., the quantile risk, naturally arrives at prediction set-based policies with a max-min decision rule.  However, VaR is not subadditive~\cite{Artzner1999}, and the resulting hard coverage threshold does not distinguish between moderate and extreme tail losses.

In this work, we formulate risk-averse decision making using OCE risk measures.  We derive the optimal policy under known distributions, and show how it reduces to a prediction set-based solution similar to ~\cite{Kiyani2025} for the CVaR. For unknown distributions, we propose a calibration algorithm that provides high-probability control of the OCE risk by leveraging learn-then-test (LTT) \cite{angelopoulos2025learn,farzaneh2026statistically}. We evaluate the approach on two wireless beamforming settings.

\section{Problem Formulation}
\label{sec:formulation}

\noindent \textbf{Setting.} Consider an agent operating on a system with true state $Y\in \mathcal{Y}$ based on an observation $X\in\mathcal{X}$, with $X$ and $Y$ modeled as random variables with joint distribution $P_{XY}$. The agent observes features $X$ and selects an action $a\in\mathcal{A}$ without observing the true state $Y$. As a result of its action $a$, the agent  receives a loss  $\ell(a,Y)$, depending on the unknown state $Y$. An action policy $a(\cdot):\mathcal{X}\to\mathcal{A}$ maps features $X$ to actions $a$. The decision maker is risk-averse: rather than minimizing expected loss, it seeks a policy whose risk-adjusted performance is controlled.

\noindent \textbf{Optimized certainty equivalent (OCE) risk.}  We measure performance via the class of  optimized certainty equivalent (OCE) risks~\cite{BenTal1986,BenTal2007}. To define this class, let $\rho:\mathbb{R}\to\mathbb{R}$ be a  convex and increasing function. For a loss random variable $L\sim P_L$, the OCE risk is
\begin{equation}\label{eq:oce}
  \OCE_\rho(L) = \min_{t\in\mathbb{R}}\big\{t + \mathbb{E}[\rho(L-t)]\big\}.
\end{equation}
The OCE risk (\ref{eq:oce}) can be interpreted as the sum of a deterministic reserve $t$ and of a random residual $L-t$ evaluated through the expected penalty $\mathbb{E}[\rho(L-t)]$. The OCE selects the best such decomposition by minimizing over the deterministic reserve $t$.

The choice of the penalty function $\rho$ determines the level of risk aversion of the decision maker. In particular, the OCE family includes many popular risk measures as special cases~\cite{BenTal2007,Gupte2026}. These encompass the mean-variance risk $\OCE_\rho(L)=\mathbb{E}[L]+\beta\Var(L)$, obtained with the penalty function $\rho(z)=z+\beta z^2$ for some $\beta\geq 0$; and the conditional value at risk (CVaR) at level $0\leq \alpha \leq 1$
\begin{equation}\label{eq:cvar}
  \OCE_\rho(L)=\CVaR_\alpha(L) = \min_{t\in\mathbb{R}}\Big\{t+\frac{1}{\alpha}\mathbb{E}\big[(L-t)^+\big]\Big\},
\end{equation}
obtained with $\rho(z)=\alpha^{-1}z^+$, where $(x)^+=\max(x,0)$. For continuous loss variables, the CVaR at level $\alpha$ equals the expected loss in the worst-$\alpha$ fraction of outcomes, i.e., $\CVaR_\alpha(L)=\mathbb{E}[L\,|\,L\geq \VaR_\alpha(L)]$, where \begin{equation}\VaR_\alpha(L)=\min\{t:\Pr(L\leq t)\geq 1- \alpha\}\end{equation} is the value-at-risk (VaR) at level $\alpha$, i.e., the $(1-\alpha)$-quantile, of the loss. The CVaR thus captures the average severity of tail losses beyond the $\VaR$ threshold.

A parametric extension of the CVaR is given by the smooth CVaR (sCVaR) $\sCVaR_{\alpha,\tau}(L) = \OCE_{\rho_{\alpha,\tau}}(L)$,
with $\rho_{\alpha,\tau}(z) = \frac{\tau}{\alpha}\ln\big(1+e^{z/\tau}\big)$,
where $\tau>0$ is the temperature parameter. The smooth CVaR recovers the standard CVaR in the zero-temperature limit, as $\lim_{\tau \to 0^+}\sCVaR_{\alpha,\tau}(L)=\CVaR_\alpha(L)$. 


\noindent \textbf{Risk-averse decision making.} Given a pre-specified penalty function $\rho$, we are interested in optimizing policy~$a(\cdot)$ with the goal of minimizing the OCE risk, i.e.,
\begin{equation}\label{eq:oce-opt}
  \min_{a(\cdot):\mathcal{X}\to\mathcal{A}}\;\OCE_\rho\big(\!\ell(a(X),Y)\big).
\end{equation}
Using \eqref{eq:cvar}, this problem can be equivalently expressed as
\begin{equation}\label{eq:oce-expanded}
  \min_{a(\cdot)}\;\min_{t\in\mathbb{R}}\;\Big\{t +
  \mathbb{E}\big[\rho\big(\!\ell(a(X),Y)-t\big)\big]\Big\}.
\end{equation}

\section{Optimal Policy for Known State-Observation Distribution}
\label{sec:optimal}

In this section, we characterize the solution to problem~\eqref{eq:oce-expanded} when the joint distribution $P_{XY}$ of observation $X$ and state $Y$ is known. We then show how the framework recovers a prediction set-based solution, in a manner similar to reference~\cite{Kiyani2025}, for the special case of the CVaR risk.

\subsection{Optimal Policy}
Since the two minimizations in (\ref{eq:oce-expanded}) are both over the same objective, they can be interchanged. Therefore, for fixed reserve $t$, the optimal action separates across covariates as
\begin{equation}\label{eq:decomp}
  \min_{a(\cdot)}\;\mathbb{E}\big[\rho(\ell(a(X),Y)\!-\!t)\big]
  = \mathbb{E}_X\Big[\min_{a\in\mathcal{A}}
  \mathbb{E}_{Y|X}\!\big[\rho(\ell(a,Y)\!-\!t)\big]\Big],
\end{equation}where the outer expectation is over the marginal $X\sim P_X$ of the observation, and the inner is over the conditional distribution $Y\sim P_{Y|X}$.

Accordingly, the optimal action for a fixed reserve $t$ is given by
\begin{equation}
  a^*(x,t) = \argmin_{a\in\mathcal{A}}\;
  \mathbb{E}_{Y|X=x}\big[\rho(\ell(a,Y)-t)\big], \label{eq:astar}
\end{equation} where the expectation is over the distribution of state $Y$ given a fixed observation $X=x$.
Therefore, the OCE risk minimization problem~\eqref{eq:oce-expanded} reduces to the scalar optimization
\begin{equation}\label{eq:scalar}
  \min_{t\in\mathbb{R}}\;\{t+\rho(\ell(a^*(X,t), Y) - t)\}.
\end{equation}

When the objective in \eqref{eq:scalar} is differentiable,  a necessary stationarity condition for the optimal reserve $t^*$ is
\begin{equation}\label{eq:stationary}
  \mathbb{E}\big[\rho'(\ell(a^*(X,t^*),Y)-t^*)\big]=1,
\end{equation}
where $\rho'(\cdot)$ denotes the first derivative of function $\rho(\cdot)$.
This is a fixed-point equation, which can be addressed as follows. First, for any fixed reserve $t$, obtain the optimal action $a^*(\cdot,t)$ from~\eqref{eq:astar}. The resulting optimality condition \eqref{eq:stationary} can then be approximately solved using methods such as the Chebyshev proxy root-finder \cite{boyd2013finding}.

\subsection{Optimal Policy via Set Prediction}

Reference \cite{Kiyani2025} studied the risk-averse decision-making problem
\begin{equation}\label{eq:radpo}
\min_{a(\cdot),\,t(\cdot)}\;\mathbb{E}[t(X)]\;\;\text{s.t.}\;\;
  \Pr\big(\ell(a(X),Y)\le t(X)\big)\ge 1\!-\!\alpha,
\end{equation}
where $t(\cdot):\mathcal{X}\to\mathbb{R}$ is a feature-dependent loss certificate. This problem is a relaxed version of the minimization of the VaR at level $\alpha$ for the \emph{conditional} distribution $P_{Y|X}$, which can be written for every fixed observation $X=x$ as \begin{equation}\min_{a,t} t \text{ s.t. } \Pr(\ell(a,Y)\leq t|X=x)\geq 1-\alpha. \end{equation} In particular, problem (\ref{eq:radpo}) can be obtained by marginalizing the constraint over the input $X$.

It is shown in \cite{Kiyani2025} that problem~\eqref{eq:radpo} can be equivalently solved by optimizing over a prediction set. Specifically, for a given action~$a$ and threshold~$t$, define the prediction set $C(x,a,t) = \{y\in\mathcal{Y}:\ell(a,y)\leq t\}$, which contains all states for which the loss does not exceed threshold~$t$. The optimal solution is obtained by jointly selecting the action $a^*(x)$ and the input-dependent threshold $t^*(x)$ such that the max-min criterion
\begin{equation}
\label{eq:best_action}
a^*(x) = \argmin_{a}\max_{y\in C(x,a,t^*(x))} \ell(a,y)
\end{equation}
is satisfied together with the marginal coverage constraint $\Pr(Y\in C(X,a^*(X),t^*(X)))\geq 1-\alpha$.

We now show that a similar  prediction set-based structure emerges naturally from our OCE framework when applied to the CVaR. Problem~\eqref{eq:radpo} targets a form of VaR, minimizing the average loss certificate $\mathbb{E}[t(X)]$ that can be guaranteed for the $(1-\alpha)$-fraction of outcomes with low loss. In contrast, the CVaR optimization \eqref{eq:oce-opt}  with OCE risk \eqref{eq:oce} minimizes the average loss over the worst $\alpha$-fraction of outcomes with high loss. In this sense, the two objectives are thus complementary, one focused on the loss on the best $(1-\alpha)$-fraction of outcomes, and the other targeting the average performance on the remaining, worse, $\alpha$-fraction of outcomes.

Consider the stationarity condition~\eqref{eq:stationary} for the smooth CVaR penalty $\rho_{\alpha,\tau}$, which takes the form
\begin{equation}\label{eq:foc-smooth}
\mathbb{E}\Big[\sigma\Big(
  \frac{\ell(a^*(X,t^*),Y)-t^*}{\tau}\Big)\Big] = 1.
\end{equation}
As $\tau\to 0^+$, the sigmoid sharpens to a step function, and the soft coverage condition~\eqref{eq:foc-smooth} becomes the hard coverage condition
\begin{equation}\label{eq:hard-coverage}
  \Pr\big(\ell(a^*(X,t^*),Y) > t^*\big) = \alpha.
\end{equation}
This means that the set $C^*(x) = \{y\in\mathcal{Y}:\ell(a^*(x,t^*),y)\leq t^*\}$ is a prediction set with marginal coverage $1-\alpha$. \new{Furthermore, taking the same limit in the optimal action \eqref{eq:astar}, the CVaR-optimal action for a fixed observation $X=x$ reduces to
\begin{equation}\label{eq:astar-cvar}
  a^*(x,t^*) = \argmin_{a\in\mathcal{A}}\;\mathbb{E}_{Y|X=x}\!\big[\big(\ell(a,Y)-t^*\big)^+\big],
\end{equation}
i.e., it minimizes the expected excess loss above the reserve $t^*$, conditional on $X=x$. Comparing \eqref{eq:astar-cvar} with the max-min rule \eqref{eq:best_action} of \cite{Kiyani2025} highlights the complementary structure of the two frameworks. Both induce a prediction-set-based action, but \eqref{eq:best_action} selects the action minimizing the maximum loss \emph{inside} the covered set $C^*(x)$, while \eqref{eq:astar-cvar} selects the action minimizing the average loss \emph{outside} it, i.e., on the tail above level $t^*$.}

Furthermore, by \eqref{eq:hard-coverage} the quantity $t^*$ corresponds to the VaR of the loss as $t^*=\VaR_\alpha(\ell(a^*(X, t^*),Y))$. Comparing this result with the average optimal objective $\mathbb{E}[t^*(X)]$ for the problem \eqref{eq:radpo} studied in \cite{Kiyani2025} one can readily show that we have the inequality 
$t^*\geq \mathbb{E}[t^*(X)]$.
 This inequality  reflects the higher level of conservatism of the CVaR objective, which focuses on the worst $\alpha$-fraction of outcomes.


\section{Data-Driven Calibration}
\label{sec:calibration}

In practice, the distribution $P_{XY}$ is unknown. In this section, we assume access to a pre-trained model $\hat{P}_{Y|X}$ providing an approximation of the  conditional distribution $P_{Y|X}$, and to a held-out calibration set $\mathcal{D}=\{(X_i,Y_i)\}_{i=1}^n$ of i.i.d.\ draws from $P_{XY}$, which is independent of any data used to train model $\hat{P}_{Y|X}$. We fix a CVaR quantile $\alpha\in(0,1]$ and a target risk level $\varepsilon\geq 0$. We seek a data-driven procedure that selects a reserve $\hat{t}$ and outputs a policy $\hat{a}(\cdot)$ guaranteeing $\CVaR_\alpha(\ell(\hat{a}(X,\hat{t}),Y))\leq\varepsilon$ with high probability. Our approach is inspired by \cite{Huang2026},  which tackled the problem of calibrating a prediction set to control the OCE risk of a scalar loss defined on the covered set via a form of LTT \cite{angelopoulos2025learn}.

For each candidate reserve $t$, we compute the model-based policy
\begin{equation}\label{eq:ahat}
  \hat{a}(x,t) = \argmin_{a\in\mathcal{A}}\;
  \mathbb{E}_{Y\sim \hat{P}_{Y|X=x}}\big[\rho(\ell(a,Y)-t)\big].
\end{equation}
Note that this policy  uses only the model $\hat{P}_{Y|X}$ and no calibration data, and that it is generally suboptimal for the true distribution $P_{XY}$.

To formally evaluate the policy, we use the calibration data with the aim of bounding the true OCE risk of the deployed policy with high probability. To this end, for each calibration point $(X_i,Y_i)$ and candidate reserve $t$, we evaluate the realized penalty on the true label $Y_i$ as
\begin{equation}\label{eq:Ri}
  \rho_i(t) = \rho\big(\ell(\hat{a}(X_i,t),\,Y_i)-t\big).
\end{equation} The true average of this quantity provides an upper bound on the OCE risk, since the OCE risk optimizes over the reserve, while the bound evaluates at the fixed value $t$, i.e.,
\begin{equation}\label{eq:rho-bound}
  \OCE_\rho\big(\ell(\hat{a}(X,t),Y)\big) \leq t+\mathbb{E}[\rho_i(t)].
\end{equation}

To construct a high-probability upper bound on the OCE risk, we introduce a finite grid $\mathcal{T}$ of candidate reserves. Assuming the penalty values are bounded, i.e., $0\leq \rho_i(t)\leq B$ for a known constant $B$, we leverage the inequality~\eqref{eq:rho-bound}, along with Hoeffding's inequality and a union bound over $\mathcal{T}$ (see, e.g., \cite{simeone2022machine}). This allows us to conclude that, with probability at least $1-\delta$ over the calibration data, the following holds simultaneously for all $t\in\mathcal{T}$:
\begin{equation}\label{eq:ucb}
  \OCE_\rho\big(\ell(\hat{a}(X,t),Y)\big) \leq
  \underbrace{t+ \frac{1}{n}\sum_{i=1}^n \rho_i(t)+B\sqrt{\frac{\log(|\mathcal{T}|/\delta)}{2n}}}_{\UCB_n(t)}.
\end{equation}

This in turn implies the following guarantee: For any candidate reserve $\hat{t}\in\mathcal{T}$ satisfying the inequality $\UCB_n(\hat{t})\leq\varepsilon$, we have the inequality
\begin{equation}\label{eq:guarantee}
  \Pr\big[\CVaR_\alpha\big(\ell(\hat{a}(X,\hat{t}),Y)\big)\leq\varepsilon\big]\geq 1-\delta,
\end{equation}
where the probability is over the calibration data. Accordingly,  the model-based policy $\hat{a}(\cdot,\hat{t})$ is certified to satisfy the condition  $\CVaR_\alpha\leq\varepsilon$ on the true distribution with confidence $1-\delta$. Algorithm~\ref{alg:oce-rcps} summarizes the calibration procedure.

\begin{algorithm}[t]
\caption{Data-Driven Calibration for OCE-Based Decision Making}
\label{alg:oce-rcps}
\begin{algorithmic}[1]
\REQUIRE CVaR quantile $\alpha$, target $\varepsilon$, confidence $1\!-\!\delta$, calibration data $\mathcal{D}=\{(X_i,Y_i)\}_{i=1}^n$, model $\hat{P}_{Y|X}$, grid $\mathcal{T}$, penalty $\rho_\alpha$, loss $\ell$.
\FOR{each $t\in\mathcal{T}$}
  \FOR{each $i=1,\ldots,n$}
    \STATE $\hat{a}(X_i,t)\leftarrow\argmin_{a\in\mathcal{A}}\;\mathbb{E}_{Y\sim \hat{P}_{Y|X=X_i}}[\rho(\ell(a,Y)-t)]$
    \STATE $\rho_i(t)\leftarrow\rho(\ell(\hat{a}(X_i,t),Y_i)-t)$
  \ENDFOR
  \STATE $\UCB_n(t)\leftarrow t+\frac{1}{n}\sum_{i=1}^n \rho_i(t)
         +B\sqrt{\log(|\mathcal{T}|/\delta)/(2n)}$
\ENDFOR
\STATE $\hat{t}\leftarrow\argmin_{t\in\mathcal{T}:\,\UCB_n(t)\le\varepsilon}\UCB_n(t)$
\ENSURE Policy $\hat{a}(\cdot,\hat{t})$ with guarantee~\eqref{eq:guarantee}.
\end{algorithmic}
\end{algorithm}

\section{Numerical Results}
\label{sec:results}

In this section, we evaluate the optimal policy \eqref{eq:astar}--\eqref{eq:scalar} and the corresponding data-driven calibration scheme in Algorithm~\ref{alg:oce-rcps} on a wireless beamforming task in two settings: a line-of-sight channel with known joint distribution $P_{XY}$ (Sec.~\ref{sec:results-gauss}), and a realistic ray-traced scene with unknown $P_{XY}$ (Sec.~\ref{sec:results-sionna}). In both settings, a base station equipped with $D\!=\!8$ antennas arranged as a uniform linear array (ULA) transmits to a single-antenna user equipment (UE); the state $Y\!\in\!\mathbb{C}^D$ is the downlink channel; the action $a \in \mathbb{C}^D$ is a normalized beamforming vector $\|a\|^2\!=\!1$; and the loss is the normalized capacity gap at some signal-to-noise ratio (SNR) level $\gamma \geq 0$, i.e.,
\begin{equation}\label{eq:loss-mismatch}
  \ell(a,Y) = 1 - \frac{\log_2\!\left(1+\gamma\,|a^\dagger Y|^2/\|h\|^2\right)}{\log_2(1+\gamma)} \in [0,1],
\end{equation}
where we set $\gamma = 10$.

\subsection{Line-of-Sight Channel}
\label{sec:results-gauss}
In this first experiment, the covariate $X$ is the estimated angle of the line-of-sight propagation from UE to BS, which is uniformly distributed in the set $[-30^\circ,+30^\circ]$, and the channel $Y$ is given by $Y\!=\!\bar{a}(X+\Delta)+n$, with $\bar{a}(\theta)$ being the array response at angle $\theta$, $\Delta$ being an angle estimation error, and $n\!\sim\!\mathcal{CN}(0,\sigma_h^2 I/D)$ with $\sigma_h\!=\!0.2$. The error $\Delta$ has a bimodal distribution supported on the set $\{-3^\circ, +3^\circ\}$. Finally, the beam $a$ is contained within the candidate family $a_\theta(X)\!=\!\bar{a}(X+\theta)$ over a fine grid of angular offsets $\theta\!\in\![-8^\circ,+8^\circ]$.

\begin{figure}[t]
  \centering
  \begin{subfigure}{\columnwidth}
    \centering
    \includegraphics[width=\columnwidth]{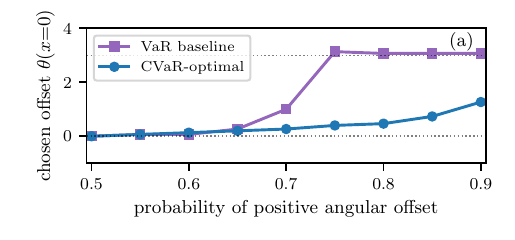}
  \end{subfigure}
  \begin{subfigure}{\columnwidth}
    \centering
    \includegraphics[width=\columnwidth]{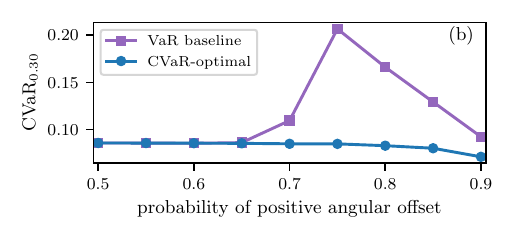}
  \end{subfigure}
  \caption{Line-of-sight channel: (a) Optimized beam steering angle at nominal angle of departure $X = 0$; and (b) $\CVaR_{0.30}$ as a function of the probability of a positive channel estimation error (equal to $+3^\circ$).}
  \label{fig:gauss}
\end{figure}

Fig.~\ref{fig:gauss} shows the optimized offset $\theta$ at $X= 0$ (Fig. \ref{fig:gauss}a) and the empirical $\CVaR_\alpha$ (Fig. \ref{fig:gauss}b) for the VaR-based policy \eqref{eq:best_action} and for the CVaR-minimizing policy \eqref{eq:astar-cvar} (implemented by using the sCVaR penalty with $\tau = 10^{-2}$ and solving equation \eqref{eq:stationary} via Chebyshev proxy method \cite{boyd2013finding}).

For an estimation error $\Delta$ with positive offset probability in the range $[0.5,0.7)$, both criteria adopt a compromise beam near $\theta\!=\!0^\circ$, yielding similar values of $\CVaR_{0.30}$. However, once the positive offset probability reaches $\!1{-}\alpha$, the VaR baseline abruptly commits to the beam offset $\theta=+3^\circ$, since worst-case outputs with probability smaller than $0.3$ do not matter when optimizing for the VaR at level $\alpha = 0.3$. This leads to a degradation in the CVaR performance. In contrast, the CVaR-optimal strategy slowly caters also to negative estimation errors as the probability of a positive estimation error increases, yielding a decreasing CVaR.

\subsection{Ray-Traced Channel}
\label{sec:results-sionna}
In this section, we study a more realistic setting in which the channel $Y$ is generated via Sionna ray tracing adopting the Munich scene \cite{Hoydis2023}. The input $X$ is the estimated UE location, and the true location is affected by a bimodal additive error with magnitude 6 m. Here, we do not assume knowledge of the joint distribution $P_{XY}$, but rather assume a model $\hat{P}_{Y|X}$ built from the same ray tracer but with electromagnetic parameters affected by an additive error with relative standard deviation $20\%$ \cite{ruah2023bayesian,ruah2026bridge}.

We set CVaR quantile $\alpha\!=\!0.20$ and target a loss no larger than $\varepsilon\!=\!0.25$ with probability at least $1-\delta = 0.95$. As the VaR baseline, we implement Algorithm~1 of~\cite{Kiyani2025} using $n\!=\!12{,}000$ calibration samples, and for the CVaR-optimal scheme we apply Algorithm~\ref{alg:oce-rcps} with the same calibration samples and $|\mathcal{T}|\!=\!30$ reserves.

\begin{figure}[t]
  \centering
  \includegraphics[width=\columnwidth]{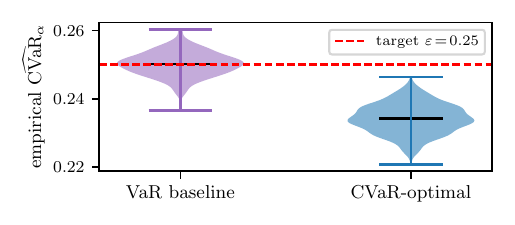}
  \caption{Ray-traced channel: Distribution of empirical $\CVaR_{0.20}$ over 500 Monte Carlo redraws of the test channel realizations, with target loss $\varepsilon\!=\!0.25$.}
  \label{fig:sionna-violin}
\end{figure}

Fig.~\ref{fig:sionna-violin} shows the distribution of empirical $\CVaR_{0.20}$ over 500 Monte Carlo redraws of the test channel realizations. The VaR baseline exhibits an empirical $\CVaR_{0.20}$ exceeding the target $\varepsilon$ in $53.4\%$ of the trials, violating guarantee \eqref{eq:guarantee} by a large margin. In contrast, the CVaR-optimal baseline satisfies the target $\CVaR_{0.20}<\varepsilon$ in all 500 trials, meeting the guarantee \eqref{eq:guarantee}.

\section{Conclusions}
\label{sec:conclusions}
In this paper, we have developed a unified framework for risk-averse decision making based on optimized certainty equivalent (OCE) risk measures. When the state--observation distribution is known, we have characterized the optimal policy through a conditional action optimization and a scalar optimization over the reserve. For CVaR, this characterization yields a prediction-set interpretation that complements VaR-based decision making. When the distribution is unknown, we proposed a data-driven calibration procedure that combines a learned conditional model with held-out calibration data to provide high-probability control of the resulting OCE risk. Numerical experiments on wireless beamforming confirmed the theoretical findings. Future work may consider larger-scale experiments and other OCE measures.

\section*{Acknowledgments}
This work was supported by the European Research Council (ERC) under the European Union's Horizon Europe Programme (grant agreement No.\ 101198347). The work of O. Simeone was also supported by an EPSRC Open Fellowship (EP/W024101/1) and by the EPSRC project (EP/X011852/1).

\balance
\bibliographystyle{IEEEbib}
\bibliography{refs}

\end{document}